# Self-supervised Representation Learning on Electronic Health Records with Graph Kernel Infomax

HAO-REN YAO, National Institutes of Health, USA
NAIREN CAO, Boston College, USA
KATINA RUSSELL, Georgetown University, USA
DER-CHEN CHANG, Georgetown University, USA
OPHIR FRIEDER, Georgetown University, USA
JEREMY T. FINEMAN, Georgetown University, USA

Learning Electronic Health Records (EHRs) representation is a preeminent yet under-discovered research topic. It benefits various clinical decision support applications, e.g., medication outcome prediction or patient similarity search. Current approaches focus on task-specific label supervision on vectorized sequential EHR, which is not applicable to large-scale unsupervised scenarios. Recently, contrastive learning has shown great success in self-supervised representation learning problems. However, complex temporality often degrades the performance. We propose Graph Kernel Infomax, a self-supervised graph kernel learning approach on the graphical representation of EHR, to overcome the previous problems. Unlike the state-of-the-art, we do not change the graph structure to construct augmented views. Instead, we use Kernel Subspace Augmentation to embed nodes into two geometrically different manifold views. The entire framework is trained by contrasting nodes and graph representations on those two manifold views through the commonly used contrastive objectives. Empirically, using publicly available benchmark EHR datasets, our approach yields performance on clinical downstream tasks that exceeds the state-of-the-art. Theoretically, the variation in distance metrics naturally creates different views as data augmentation without changing graph structures. Practically, our method is non-ad hoc and confirms superior performance on commonly used graph benchmark datasets.

## 1 INTRODUCTION

Representation learning of patient Electronic Health Records (EHRs) is the foundation for data-driven personalized healthcare and clinical decision support [39]. Many approaches, in particular deep learning models, were proposed to learn EHR representation [4, 5, 24, 38, 44]. However, all of them rely on task-specific label supervision. Annotation on

Authors' addresses: Hao-Ren Yao, hao-ren.yao@nih.gov, National Institutes of Health, Bethesda, USA; Nairen Cao, caonc@bc.edu, Boston College, USA; Katina Russell, ker83@georgetown.edu, Georgetown University, USA; Der-Chen Chang, chang@georgetown.edu, Georgetown University, USA; Ophir Frieder, ophir@ir.cs.georgetown.edu, Georgetown University, USA; Jeremy T. Fineman, jfineman@cs.georgetown.edu, Georgetown University, USA.

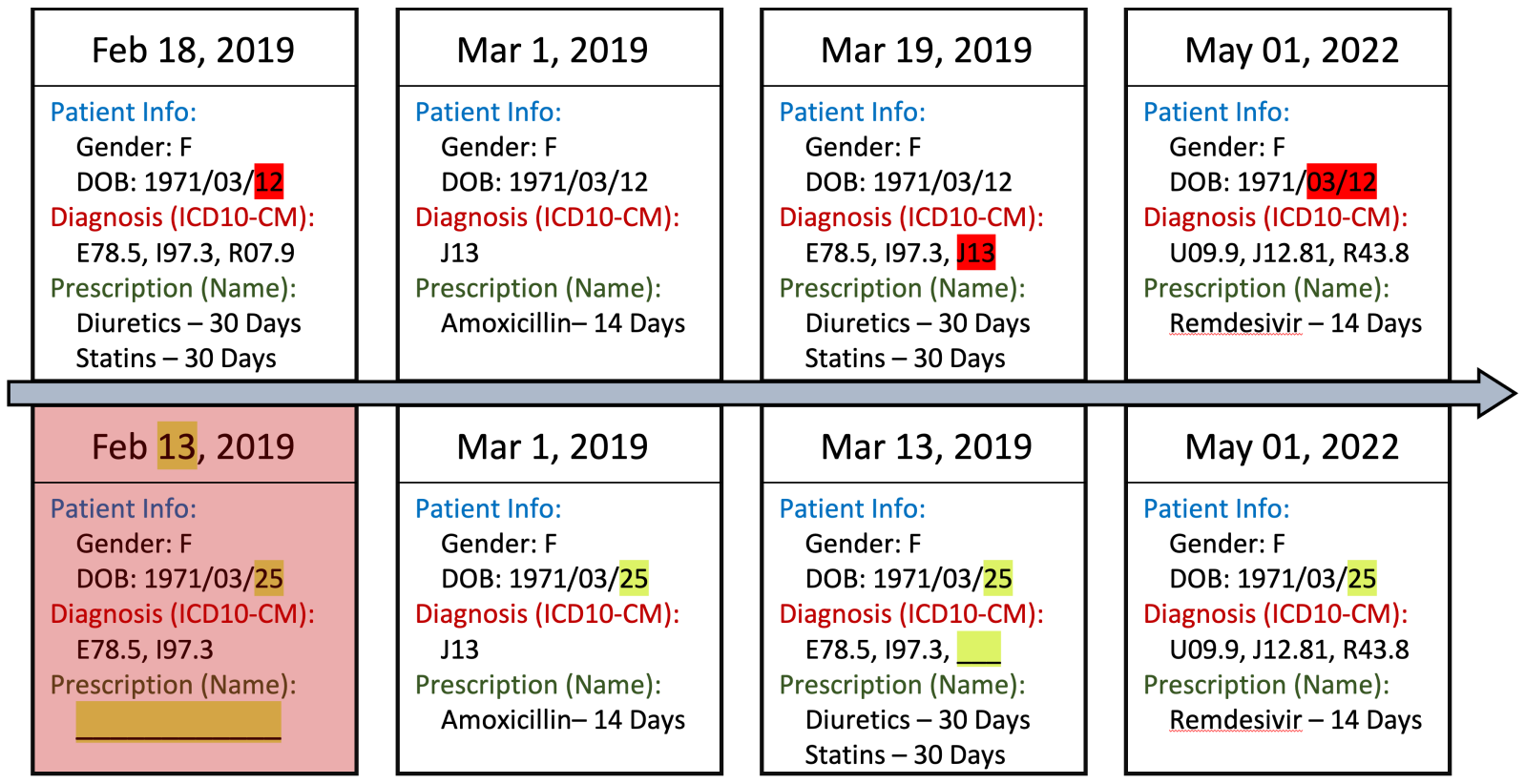

Fig. 1. An example of view ambiguity. Two closely matched patients, denoted as A (above) and B (below), each falling into distinct target cohorts, e.g., individuals with unsuccessful hypertension treatment outcomes versus those with successful treatment plans. When applying data augmentation, highlighted in red, such as historical data cropping (which entails the removal of the initial clinical visit for patient B) and the introduction of Gaussian noise (involving masking out the date of DOB and diagnosis code for patient A), the disparities between these two patients are mitigated and potentially increases their overall similarity. However, this effect also introduces ambiguity into the augmented views of patients A and B, which, in turn, may lead to biased contrastive learning.

large-scale EHR is challenging and, thus, not ideal for an unknown clinical problem with unlabeled medical records, e.g., patient similarity search [28]. Autoencoders [23, 26] provide the option for unsupervised learning. Regardless, sparse medical coding and lengthy temporal sequences often lead to poor reconstruction loss optimization. Though the BERT-based language model [22] shows remarkable clinical document pre-training, it still requires downstream clinical task fine-tuning to obtain promising performance. Yet, a universal representation is still lacking.

On the other hand, contrastive learning has shown state-of-the-art performance in a fully unsupervised setting, e.g., without fine-tuning, in computer vision [3] and graph representation learning [36]. It has been applied to EHR representation learning and achieves superior effort in several clinical downstream tasks [45]. However, data augmentation on the long-time span of high variance and temporality of EHR often causes view ambiguity: the similarity increases on negative samples but decreases on the positive ones. As in Figure 1, two highly similar patients but reside in different target cohort groups become even more similar after random history crop and Gaussian noise [45]. Thus, designing a proper semantic-preserving data augmentation without modification to the original data with such EHR complexity is necessary to overcome these challenges.

Addressing these challenges, we design a theoretically and empirically sound self-supervised, contrastive graph kernel learning approach, namely, ***Graph Kernel Infomax*** **(GKI)**. Compressing feature temporality and heterogeneity of EHR via graph, we learn graph kernel feature maps with a graphical representation of EHR to model universal patient representation. Instead of modifying graph topologies for augmentation, we propose Kernel Subspace Augmentation to embed nodes into two manifolds yielded by different kernels with different underlying metrics. The embedding of node and graph on two manifolds act as augmented views for commonly used contrastive loss. We geometrically analyze data augmentation under a local-neighborhood perspective and reveal that adjusting distance metrics in the underlying manifold can achieve the same effect as the augmentation on graph structure. By maximizing Mutual Information (MI) between node and graph in these two manifolds, our patient representation can directly apply to linear models on clinical tasks without fine-tuning and label supervision. To our best knowledge, this is the first effort for self-supervised representation of EHR without any task-specific supervision. Furthermore, our method achieves superior or at least competitive performance on other graph benchmarks, revealing its non-ad hoc generalizability and robustness to handle non-clinical domains. Our contributions are as follows:

- We propose Graph Kernel Infomax for self-supervised representation learning on EHR without the need to modify graph structures for data augmentation.
- We develop and provide theoretical analysis on Kernel Subspace Augmentation by embedding nodes and graph to geometrically different manifolds via a kernel method to construct augmented views.
- We experiment with a large-scale, publicly available EHR dataset to demonstrate our effectiveness on clinical downstream tasks.
- We validate our non-ad hoc generalizability and robustness with widely used graph benchmark datasets.

## 2 RELATED WORK

### 2.1 Patient Representation Learning

Many efforts focus on EHR representation learning. These range from traditional electronic phenotyping [39] to deep representation learning on end-to-end temporal clinical event modeling with task-specific label supervision, particularly, attention-based RNN model [4, 24]. Another line of research focuses on hierarchical temporality by graph-based EHR [5, 42–44]. Besides the autoencoder-based model [26], there is less work on unsupervised patient representation learning. One reason is the non-trivial design of training signals without label supervision for complex EHR [31]. The evolution of self-supervised pre-trained models [6], trained on large-scale EHR, shows promising results on various clinical downstream tasks [22]. The recently proposed contrastive learning method [3] shines a new light on learning effective representation without task-specific fine-tuning. It further demonstrates the utility of medical representation learning [45]. Nonetheless, current work [45] only applies to time-series of Intensive Care Units (ICU) and EEG. Yet, the unsupervised EHR representation learning is not fully solved.

### 2.2 Graph Representation Learning

Graph Contrastive Learning has become a mainstream learning framework to improve unsupervised representation learning tasks on graph datasets. It was first proposed in Deep Graph Infomax (DGI) [36], in which the Infomax principle was introduced for node representation learning by maximizing the mutual information between graph local and global contexts. InfoGraph [32] extends such an idea on learning graph-level representation with various granularity, followed by GCC [30], which evaluates the mutual information between different local sub-graphs extracted by random walks. Recently, GraphCL [46] and MVGRL [13] have introduced several augmentation strategies to construct different views for contrastive learning and achieve the state-of-the-art. We can consider such augmentation schemes, namely corrupting node and graph information or forming new topological properties on graph structures, and try to create perturbation to let the model learn view-invariant feature information. However, such view creation may pose resource overhead on GPU memory or construct less informative views due to data complexity, where our evaluation confirms the difficulty of clinical application.

### 2.3 Manifold Learning with Kernel

A manifold (e.g., Riemannian manifold) is a topological space that locally approximates Euclidean space. Usually, it models complex non-linear data that resides on spheres or hyperbolic spaces where Euclidean distance is no longer applicable. It is well-known that Symmetric Positive Definite (SPD) matrices form a Riemannian manifold [7, 15]. However, directly applying Euclidean-based learning algorithms results in undesirable effects [7] since the Euclidean distance only resembles local structure instead of manifold non-linearity geometry. Several methods were proposed to approximate SPD manifold [7], for instance, the kernel method, which embeds a Riemannian manifold into a high dimensional Reproducing Kernel Hilbert Space (RKHS) [15] via positive definite kernel, where inner product is naturally applied. In [49], a spherical manifold is approximated by Gaussian Kernel for unsupervised domain adaptation. The distance metric, defining a kernel function, is also investigated in [2, 9], which leads to the generalization on certain shapes of a manifold, e.g., geodesic Laplacian kernel for a spherical manifold.

## 3 METHODOLOGY

In this section, we present Graph Kernel Infomax (GKI), a self-supervised contrastive learning framework for learning kernel feature maps for the graphical representation of EHR. Figure 2 illustrates the architecture of GKI.

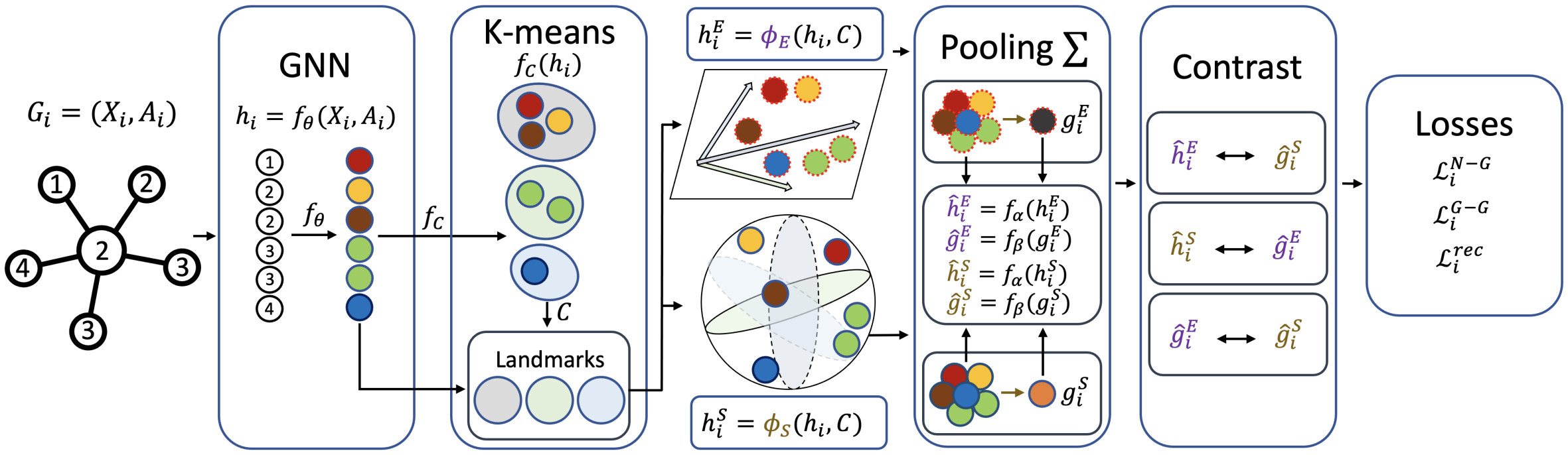

Fig. 2. Graph Kernel Infomax. We first obtain node embeddings from the GNN $f_\theta$ followed by K-means clustering $f_C$. The resulting cluster centroids $C$ are used as landmark points for the nyström kernel approximation to construct kernel feature maps $\phi_E$ and $\phi_S$ for Euclidean and spherical distance kernels. The two kernel embeddings for nodes and graph are contrasted to compute final losses.

### 3.1 Problem Definition

Let $G_s = \{G_1, G_2, \ldots\}$ be a set of graphs (EHR graphs) and $G_i \in G_s$ denotes $i$-th patient EHR graph. For each $G_i = (X_i, A_i)$ containing $n$ nodes, we denote the node feature matrix and adjacency matrix as $X_i \in \mathbb{R}^{n \times D}$ and $A_i \in \mathbb{R}^{n \times n}$, where each node is a $D$-dimensional feature vector in $X_i$, and an edge connection between two nodes is either binary or weighted in $A_i$. Our goal is to learn graph-level kernel feature maps $\Phi_E, \Phi_S$ derived from Euclidean distance and spherical distance kernels without label supervision via contrastive learning. The resulting graph-level kernel embeddings $\Phi_E(G_i)$ and $\Phi_S(G_i)$, acting as patient representation under Euclidean and spherical distance kernels, can be directly supplied to downstream task models, e.g., linear SVM. It also benefits from the kernel method and multiple kernel properties from two kernels providing efficient non-linear learning for linear classifiers and similarity search on large-scale EHR. We summarize Graph Kernel Infomax in Figure 2.

### 3.2 EHR Graphical Representation

We formulate a patient's EHR, Figure 3, as a directed graph following the definition in [44], except we do not impose directed-acyclic properties. Every medical event, including a disease diagnosis, a drug prescription, and other medical codes, is represented as a node with one-hot encoding. The edge between two nodes represents an ordered direct connection with the time difference as an edge weight (e.g., days or prescription period). Note that we assume the duration for an event is the prescription period, and any medical event in between will create another branch for connection. The demographic information of the patient, e.g., gender, connects to the first medical event with age as an edge weight. Figure 4 shows an example of a graphical representation of an EHR, and Figure 5 is a sample patient graph built on the MIMIC-IV dataset. As we can see, we retain overlaps between medical events instead of connecting sequentially to preserve their original temporal structure. The construction steps and higher-resolution sample patient graphs from the MIMIC-IV dataset are in the Appendix.

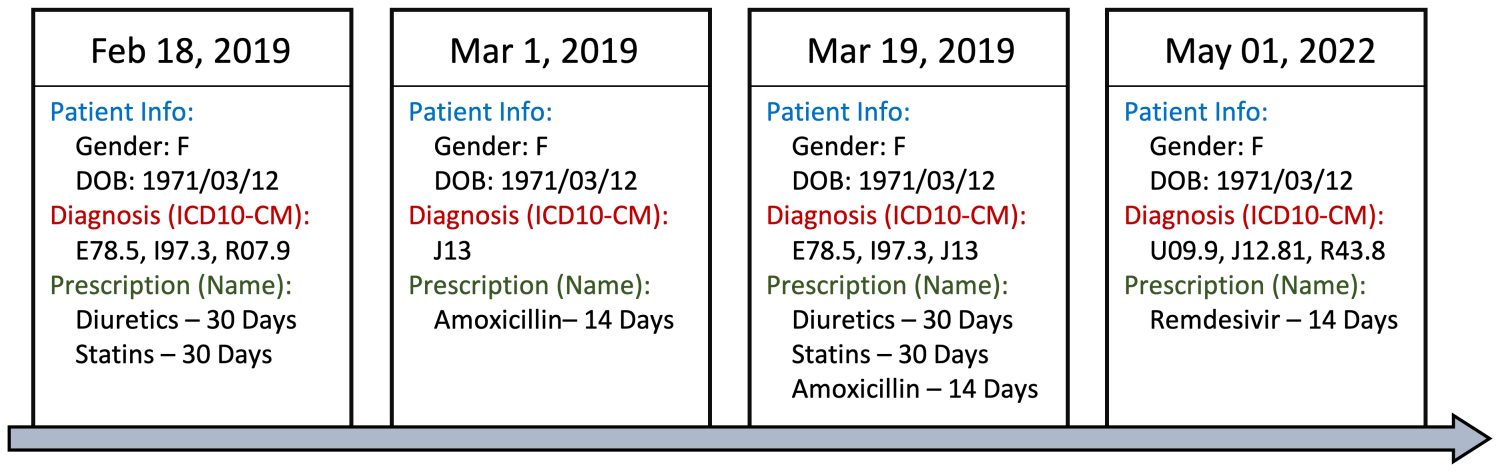

Fig. 3. A sample patient EHRs.

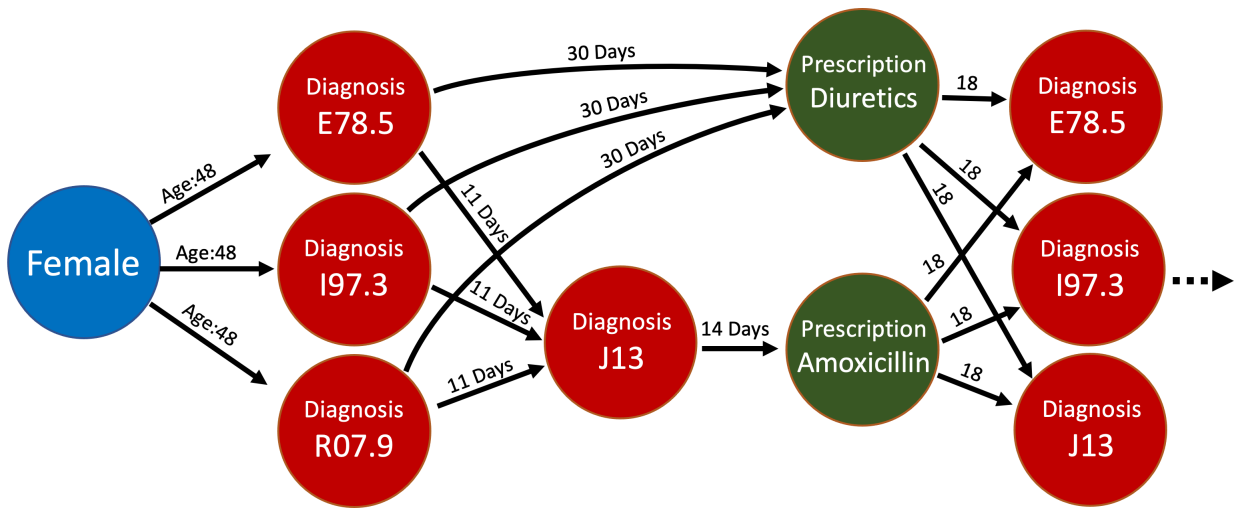

Fig. 4. An example of a patient graph reflecting the first three medical events in Figure 3.

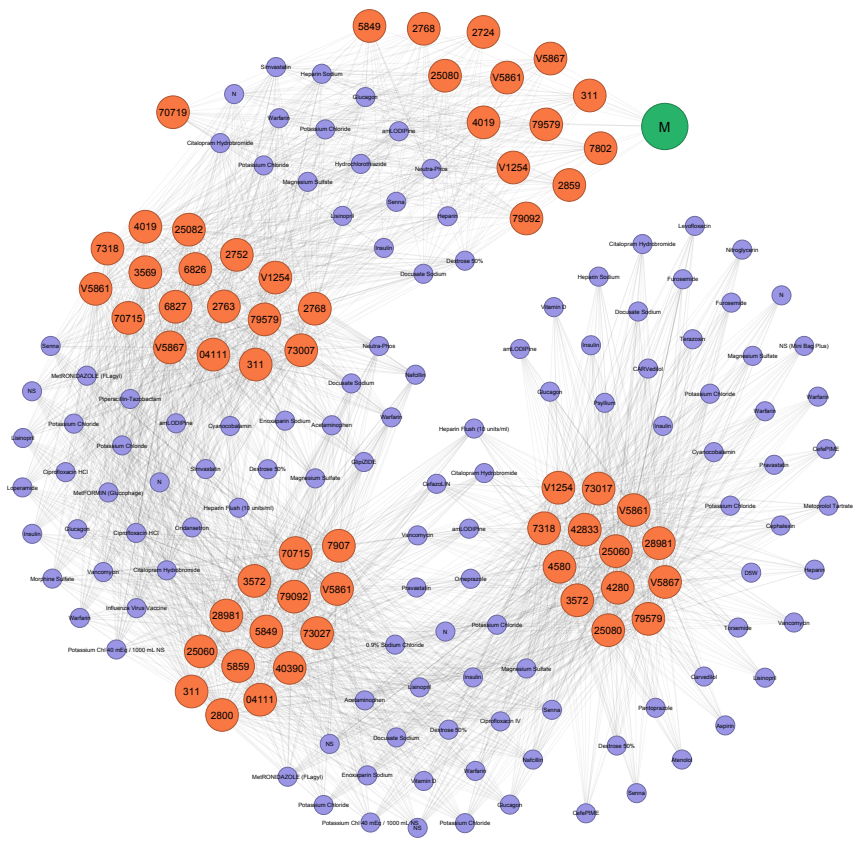

Fig. 5. A sample patient graph constructed from the MIMIC-IV dataset. Higher-resolution images can be found in the Appendix.

### 3.3 Graph Kernel Feature Map

We construct graph-level kernel feature map $\Phi$ given with a RKHS kernel $k$ by the following components: **(1)** graph encoder, **(2)** K-means clustering, and **(3)** Node-level feature map via Nyström kernel approximation [8].

*3.3.1 Graph Representation Learning.* Given a patient graph $G_i = (X_i, A_i)$, we derive its $l$-th layer $d$-dimensional node representation via graph neural networks (GNNs) as:

$$h_i^{(l)} = f_\theta(X_i, A_i) \in \mathbb{R}^{n \times d} \tag{1}$$

where $f_\theta$ denotes a GNN layer, and we allow other choices of GNN architecture.

*3.3.2 K-Means Clustering.* Nyström method avoids computing the full-rank kernel matrix by landmark selection, which is critical to approximation quality. We adopt K-means clustering, which has proven to generate a good kernel approximation [47], and select cluster centroids as landmarks. To enable end-to-end mini-batch training, we develop a $l$-th layer K-means clustering $f_{C^{(l)}}$ to partitions $h_i^{(l)}$ into $K$ clusters:

$$H_i^{(l)} = f_{C^{(l)}}(h_i^{(l)}) = \sigma(h_i^{(l)} {C^{(l)}}^T) \tag{2}$$

where $C^{(l)} \in \mathbb{R}^{K \times d}$ denotes $l$-th layer $K$ randomly initialized cluster centroids, $\sigma(\cdot)$ is the sparse-softmax activation function [25] and $H_i^{(l)} \in \mathbb{R}^{n \times K}$ is the clustering assignment for $h_i^{(l)}$. The associated clustering loss for $h_i^{(l)}$ is defined as:

$$\mathcal{L}_i^{\text{rec}^{(l)}} = \|h_i^{(l)} - H_i^{(l)} C_i^{(l)}\|_F \tag{3}$$

where $\|\cdot\|_F$ denotes frobenius norm. The final clustering loss is computed from all layers as $\mathcal{L}_i^{\text{rec}} = \sum_{l=1}^{L} \mathcal{L}_i^{\text{rec}^{(l)}}$.

*3.3.3 Graph-level Kernel Feature Map.* To construct graph-level kernel feature map, we first define $l$-th layer node-level kernel feature map for $h_i^{(l)}$ given $C^{(l)} = \{c_1^{(l)}, \cdots, c_K^{(l)}\}$ selected as landmark points as:

$$\phi(h_i^{(l)}) = [k(h_i^{(l)}, c_1^{(l)}), \dots, k(h_i^{(l)}, c_K^{(l)})]\hat{\kappa}^\dagger \in \mathbb{R}^{n \times K} \tag{4}$$

where $\hat{\kappa}_{ij} = k(c_i, c_j)$ is the pseudo-inverses of $\mathcal{K} \in \mathbb{R}^{K \times K}$ where $\mathcal{K}_{ij} = k(c_i, c_j)$. One interesting observation is that, by setting the pseudo-inverse $\hat{\kappa}^\dagger$ to $I$, we found a dramatic performance improvement in our evaluation. We believe such an effect leads to orthogonal constraints on landmark points that act as orthogonal regularizations to avoid trivial solutions in K-means clustering. We leave it here for future research work. We then derive the graph-level kernel feature map for the entire graph. In order to preserve high-order information on the graph, we concatenate the summation on the node-level kernel feature map in each layer:

$$\Phi(G_i) = \Big\|_{l=1}^{L} \left[ \sum_{j=1}^{n} \phi(h_i^{(l)})_j \right] \in \mathbb{R}^{KL} \tag{5}$$

### 3.4 Kernel Subspace Augmentation

The graph kernel feature map $\Phi$ is parameterized by GNN layers and K-Means clustering. To learn optimal model parameters without label supervision, we formulate it as graph contrastive learning. To design an augmentation scheme without modification to the original graph, we seek another interpretation of contrastive learning:

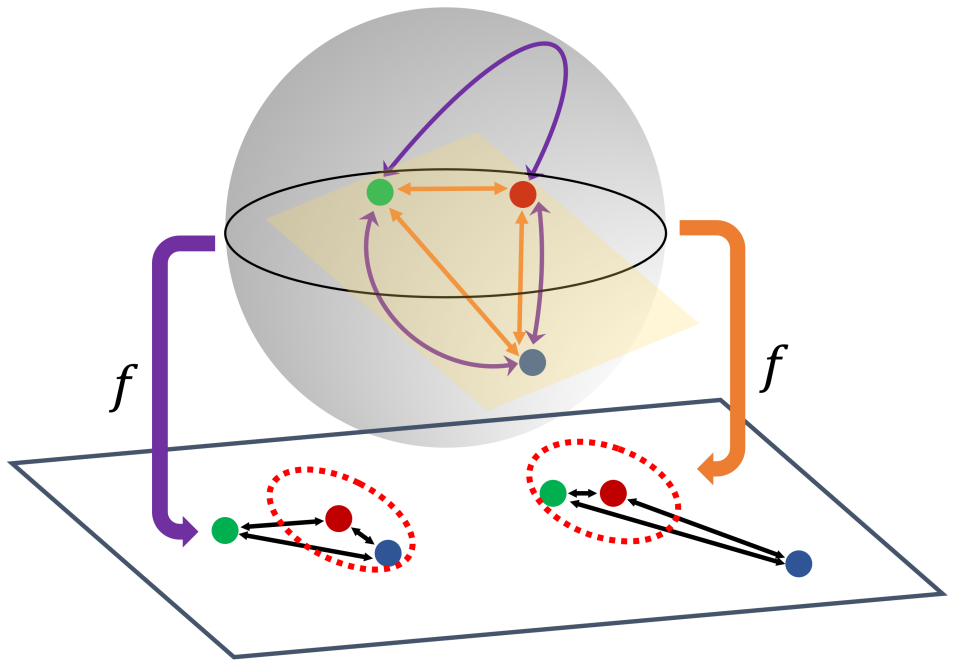

Fig. 6. Neighborhood difference. The green, red, and blue three points are measured with geodesic distance (purple) and Euclidean distance (orange). After projecting to the shared space via $f$ from their source manifold, we can see the nearest neighbor for the red point is changed.

Observation 1. *[20] Given a positive pair $z_i, z_j$, minimizing the contrastive loss for $z_i, z_j$ is equivalent to maximizing the probability that $z_j$, the positive pair of $z_i$, is the 1-nearest neighbor (1-NN) of $z_i$. Therefore, contrastive Learning can be considered as optimizing the 1-Nearest Neighbor problem.*

*3.4.1 Data Augmentation via Distance Metrics.* Observation 1 provides an intuitive explanation of explicit augmentation, which adds negative samples to create perturbation in the local neighborhood for each sample. Such perturbation alters 1-NN by adding noise in the local proximity of positive pairs in their projection space, providing signals for contrastive loss to minimize. Considering a smooth compact $d$-dimensional Riemannian manifold $\mathcal{M}$ and its true geodesic distance $d_{\mathcal{M}}$ [21], We define two distance metrics on $\mathcal{M}$, namely Euclidean and spherical distance, to approximate the true underlying geodesic distance $d_M$:

Definition 1. ***Euclidean Distance***. *The Euclidean distance $d_E$ between two arbitrary points $x_i$ and $x_j$ on $\mathcal{M}$ is defined as*

$$d_E(x_i, x_j) = \|x_i - x_j\|_2. \tag{6}$$

Definition 2. ***Spherical Distance***. *Let $S$ be a $d$-dimensional $r$-radius sphere centered at $c$ on $\mathcal{M}$ and $\pi$ be the orthogonal projection from the $\mathcal{M}$ to $S$, the spherical distance between two arbitrary points $x_i$ and $x_j$ on $\mathcal{M}$ is defined as*

$$d_S(x_i, x_j) = r\arccos(\frac{\pi(x_i) - c}{r} \cdot \frac{\pi(x_j) - c}{r}). \tag{7}$$

Theorem 1. **Distance Variation Bound**. Given two arbitrary sufficiently close neighbors $x_i$ and $x_j$ on $\mathcal{M}$, the following inequality holds when the true underlying geodesic distance between $x_i$ and $x_j$ is $m$ such that $d_{\mathcal{M}}(x_i, x_j) = m$,

$$-O(m^4) \leq d_S(x_i, x_j) - d_E(x_i, x_j) \leq O(m^3). \tag{8}$$

We prove Theorem 1 by extending the results from [21]:

Proof. Given two arbitrary neighbors $x_i$ and $x_j$ on $\mathcal{M}$, we define the Euclidean distance approximation error as $d_E(x_i, x_j) - d_{\mathcal{M}}(x_i, x_j)$ and the spherical distance approximation error as $d_S(x_i, x_j) - d_{\mathcal{M}}(x_i, x_j)$. Let $d_{\mathcal{M}}(x_i, x_j) = m$, according to [21], the distance approximation error bounds of $d_E(x_i, x_j)$ and $d_S(x_i, x_j)$ are

$$m - O(m^3) \leq d_E(x_i, x_j) \leq m \tag{9}$$

$$m - O(m^4) \leq d_S(x_i, x_j) \leq m \tag{10}$$

Multiplying -1 to Equation 10:

$$-m \leq -d_S(x_i, x_j) \leq -m + O(m^4) \tag{11}$$

Adding Equation 9 and Equation 11:

$$m - O(m^3) - m \leq d_E(x_i, x_j) - d_S(x_i, x_j) \leq m - m + O(m^4) \tag{12}$$

$$-O(m^3) \leq d_E(x_i, x_j) - d_S(x_i, x_j) \leq O(m^4) \tag{13}$$

$$-O(m^4) \leq d_S(x_i, x_j) - d_E(x_i, x_j) \leq O(m^3) \tag{14}$$

which completes the proof. □

By Theorem 1, we show that perturbation can be created by varying distance metrics to affect the distance calculation between two arbitrary points in their local-neighborhood. Such a perturbation alters 1-NN for each sample on two different manifolds. Namely, the positive pair for each sample now becomes the same data point but the embeddings on two manifolds, respectively. As illustrated in Figure 6, when projecting them to the space where contrastive loss is applied, the 1-NN for each sample will not be identical due to the perturbation on their local-neighborhood distance calculation. This inconsistency acts as a contrastive signal to optimize the model. The error between two distance measures in local-neighborhood is also bounded as shown in Theorem 1 resulting in stable training since the model will not overfit on a certain distance metric. Empirically, our model can converge to a lower minimum as described in Figure 9.

*3.4.2 Manifolds as Augmented Views.* Referring to related work, a positive definite kernel embeds a Riemannian manifold into RKHS [15]. Moreover, Nyström kernel approximation aims to find a subspace $\mathcal{H}'$, which spanned by landmark points $C$, under kernel induced RKHS [34]. Hence, we can interpret Nyström kernel feature maps as an efficient mapping that embeds input to its respective Riemannian manifolds. Applying Theorem 1, we are now finding two Riemannian manifolds with different underlying metrics that act as augmented views. As a result, we define two positive definite kernels with Euclidean and spherical distance:

Definition 3. ***Euclidean and Spherical Kernels***. *Given two samples $x$ and $y$, the Euclidean kernel $k_E$ and spherical kernel $k_S$ are defined as*

$$\begin{aligned} k_E(x, y) &= \exp(-d_E(x, y)), \\ k_S(x, y) &= \exp(-d_S(x, y)). \end{aligned} \tag{15}$$

It is already known that $k_E$, the Euclidean kernel, is a positive definite kernel. For $k_S$, we can refer [9] such that the Laplacian kernel, as the formulation in Definition 3, is positive definite when geodesic distance is used. Since we have two RKHS kernels, we can define their node and graph kernel feature maps under Euclidean kernel as $\Phi_E, \phi_E$ and spherical kernel as $\Phi_S, \phi_S$, respectively. Given a patient graph $G_i = (X_i, A_i)$ and last layer node representation $h_i$, we now define the augmented views as:

$$h_i^E = \phi_E(h_i) \tag{16}$$

$$h_i^S = \phi_S(h_i) \tag{17}$$

$$g_i^E = \Phi_E(G_i) \tag{18}$$

$$g_i^E = \Phi_S(G_i) \tag{19}$$

where $h_i^E, h_i^S$ denote node Euclidean and spherical views and $g_i^E, g_i^S$ denotes graph Euclidean and spherical view, respectively.

### 3.5 Training Graph Kernel Infomax

We now formally present Graph Kernel Infomax (GKI) and its training procedure. GKI maximizes MI by contrasting the node Euclidean view and graph Spherical view and vice versa. As we discovered, some patients contain a long history in their EHR resulting in larger graphs, while others are small. Such variation in graph size with EHR augmentation ambiguity easily makes the model overfit on certain local structures or global representations. To overcome such complex EHR problems, we contrast both node-graph and graph-graph representations.

We randomly sample size $N$ mini-batch $\mathcal{B}$ from $G_s$ for training. We train Graph Kernel Infomax via normalized temperature-scaled cross-entropy loss (NT-Xent) [3]. For positive pair $p, q$, we define the contrastive loss:

$$\ell_{cl}(p,q) = -log \frac{e^{\text{sim}(p,q)/\tau}}{\sum_{q\prime \in \mathcal{B}} \mathbb{1}_{[p \neq q\prime]} e^{\text{sim}(p,q\prime)/\tau}} \tag{20}$$

where $\text{sim}(\cdot)$ denotes cosine similarity and $\mathbb{1}_{[p \neq q\prime]} \in \{0,1\}$ defines an indicator function. Under correlated views for $G_i \in \mathcal{B}$, its node Euclidean and spherical views $h_i^E$, $h_i^S$, and graph Euclidean and spherical views $g_i^E$, $g_i^S$ are projected to a shared embedding space $\mathbb{R}^{d_p}$ by node projection head $f_\alpha$ and graph projection head $f_\beta$, producing projected nodes $\hat{h}_i^E \in \mathbb{R}^{n \times d_p}$, $\hat{h}_i^S \in \mathbb{R}^{n \times d_p}$ and projected graphs $\hat{g}_i^E \in \mathbb{R}^{d_p}$, $\hat{g}_i^S \in \mathbb{R}^{d_p}$. The projection heads are discarded after training. We define node-graph contrastive loss $\mathcal{L}_i{}^{\text{N-G}}$ and graph-graph contrastive loss $\mathcal{L}_i{}^{\text{G-G}}$ for $G_i \in \mathcal{B}$ as follows:

$$\mathcal{L}_i{}^{\text{N-G}} = \sum_{\hat{z}_E \in \hat{h}_i^E} \sum_{\hat{z}_S \in \hat{h}_i^S} \ell_{cl}(\hat{z}_E, \hat{g}_i^S) + \ell_{cl}(\hat{z}_S, \hat{g}_i^E) \tag{21}$$

$$\mathcal{L}_i{}^{\text{G-G}} = \ell_{cl}(\hat{g}_i^E, \hat{g}_i^S) + \ell_{cl}(\hat{g}_i^S, \hat{g}_i^E) \tag{22}$$

The final loss $\mathcal{L}$ is the summation of all losses defined so far for all graphs in $B$:

$$\mathcal{L} = \sum_{i=1}^{N} \mathcal{L}_i{}^{\text{N-G}} + \mathcal{L}_i{}^{\text{G-G}} + \mathcal{L}_i{}^{\text{rec}} \tag{23}$$

We optimize Graph Kernel Infomax by using mini-batch Stochastic Gradient Descent to minimize mini-batch loss $\mathcal{L}$. Note, the clustering loss is optimized jointly with contrastive loss to encourage both graph encoder and clustering to learn a clustering-friendly representation as in [40].

It should be noticed that our method is different from the parallel work [41][1], although they perform similar graph contrastive learning between Euclidean and hyperbolic space embeddings. First, we are completely unsupervised, while they require label supervision. Second, we create perturbation from distance metrics with theoretical justification, however, it is unclear whether hyperbolic space achieves similar outcomes. Third, we approximate manifolds via kernel feature maps, which directly apply to downstream linear models instead of neural space transformation with end-to-end training. In this case, we decouple pre-training and downstream tasks to achieve more generalizability.

[1]We do not include [41] in our experiments due to the lack of source codes and different evaluation settings in their original paper.

## 4 EXPERIMENTS

We perform an extensive evaluation on Graph Kernel Infomax to validate the following questions:

- **Q1 Clinical evaluations.** Does our method outperform state-of-the-art without supervised pre-training and task-specific fine-tuning on each disease across different tasks?
- **Q2 Robustness of augmentation.** Does our proposed augmentation scheme work with different graph encoders?
- **Q3 Ad hoc versus non-ad hoc.** Does the proposed framework provide non-ad hoc generalizability on non-clinical domains?

### 4.1 Experimental Setup

*4.1.1 Dataset.* For clinical evaluation, we use the MIMIC-IV dataset [16] on PhysioNet, a large-scale publicly available Electronic Health Record, with over 60,000 de-identified patients from 2008 to 2019. As confirmed by our collaborated clinician, MIMIC-IV contains sufficient long-time span clinical visits that best evaluate pre-training performance under highly variant EHR temporality. As discussed with medical experts, we evaluate our model on two real-world clinical tasks: **(i) Treatment plan mortality prediction:** predict if the given treatment plan controls chronic disease progression to avoid patient mortality. Success and failure cases are determined by whether the mortality mark exists in the last clinical visit, **(ii) Patient similarity search:** search for the most similar patient in the patient database. We select three chronic diseases, e.g., hypertension, hyperlipidemia, and diabetes, as advised by clinicians, for their well-defined diagnosis and treatment guidelines [1, 11, 33]. Patients with at least one target disease as the primary diagnosis are selected[2].

To demonstrate the non-ad hoc generalizability of our method to other non-clinical domains, we evaluate GKI on widely used graph classification benchmark TUDataset [27] that ranges from bioinformatics to social network domains. As this benchmark represents a standard evaluation tool for majority supervised and self-supervised graph representation tasks, we select eight commonly used datasets in related works [12, 13, 30, 32, 37, 46] including PTC-MR, PROTEINS, NCI1, COLLAB, IMDB-BINARY, IMDB-MULTI, REDDIT-BINARY (RDT-B), and REDDIT-MULTI-5K (RDT-M5K). The dataset statistics for both MIMIC-IV and graph classification benchmark datasets are in the Appendix.

*4.1.2 Evaluation protocol.* We evaluate GKI under linear evaluation protocol [3]. The unsupervised pre-training is firstly performed on all patient EHRs in the MIMIC-IV dataset with all selected chronic disease patients excluded. The learned representations from the frozen model are directly applied without supervised fine-tuning and evaluated with 5 times repeated 10-fold cross-validation to clinical downstream tasks. As most related works only perform one-time 10-fold cross-validation or fixed train, validation, and test sets split, we believe repeating 10-fold cross-validation 5 times is sufficient to achieve a trade-off between computing time and reliable evaluation, specifically, given a large number of evaluation runs for performance comparison, hyperparameters tuning, and model analysis. We formulate Treatment plan mortality prediction as a binary classification problem with linear logistic regression and patient similarity search as a top-K similarity search by the inner product via K-nearest neighbors where the ground-truth labels are their primary diagnosed disease. For graph classification, we follow the identical linear evaluation protocol in [13]. The frozen features are evaluated on linear SVM with 5 times repeated 10-fold cross-validation.

[2] We say a chronic disease is primarily diagnosed if such the one is the first diagnosis for a given patient's admission history.

### 4.2 Baselines

We select four types of baselines in which the source codes are publicly available: **Supervised Learning**, **Unsupervised Learning**, **Contrastive Learning**, and **Graph Contrastive Learning**. For Supervised Learning, **Retain** [4] and **Dipole** [24] are the selected RNN-based EHR representation learning methods. **GCT** [5] is also chosen as the transformer graph-based baseline. For Unsupervised Learning, we select **DeepPatient** [26], **Seq2SeqAE** [23], and **BEHRT** [22]. The BEHRT is the well-known BERT-based pre-trained model on structured EHR datasets. The pre-training is unsupervised, however, we fine-tune with label supervision on downstream clinical tasks [3]. For contrastive learning, we choose **SimCLR** [3], **SCL** [17] and **NCL** [4] [45]. We employ the identical data augmentation procedure for the EHR dataset in NCL. Finally, we choose four state-of-the-art graph contrastive learning methods: **GCC** [30], **InfoGraph** [32], **GraphCL** [46], and **MVGRL** [13] are selected as the baselines. Note, **GCC** is only reported in the graph classification benchmarks due to the OOM (Out of Memory) error during pre-training on MIMIC-IV. In addition to SOTA frameworks, we also pick **Logistic Regression**[5], **Random Forest**, and **Gradient-boosted trees** to serve classic baselines.

All end-to-end training and fine-tuning models concatenate with a single MLP for classification. SCL and NCL are evaluated by linear evaluation protocol with supervised pre-training on the training set and evaluated with linear logistic regression on frozen features. For classic baselines, the input is formulated as a bag-of-word frequency count vector on all medical codes from whole patient EHRs.

### 4.3 Implementation details

For clinical downstream tasks, we adopt a two-layer 128 hidden dimension GNN taken from GCN [19], GAT [35], and GIN [37] [6] as a graph encoder with PReLU activation function [14] for each GNN layer. For both node and graph projection head $f_\alpha$ and $f_\beta$, we employ three-layer MLP with PReLU activation function. For spherical distance, we assume it is on a 1-radius sphere centered at the origin, e.g., setting $r = 1$ and $c = 0$. As two types of patient representation from Euclidean and spherical graph-level kernel embeddings, we concatenate them into the final patient representation.

To fairly compare our model in clinical downstream tasks with other approaches, we fix the hidden dimension to 128 for all baselines and adopt the same graph encoder setting for baselines containing the GNN layer. Other configurations for all baselines are followed with their original paper implementation. The $k$-means clustering is implemented by one-layer MLP with $k$ selected from [16, 32, 64, 128, 256, 512]. The temperature parameter is fixed as 0.01 with cosine similarity. The C parameter is chosen from $[50^{-4}, 50^{-3}, \ldots, 50^{3}, 50^{4}]$ for linear logistic regression. Finally, the batch size, epochs, and learning rate for the Adam [18] optimizer are fixed for all models to 128, 100, and 0.001 accordingly. In graph classification benchmarks, we adopt the identical model configuration and parameter search in [13] with 512 hidden dimension GCN as graph encoder[7]. The number of clusters is fixed to 128. The C parameter is chosen from $[10^{-4}, 10^{-3}, \ldots, 10^{3}, 10^{4}]$ for linear SVM. Detailed hyperparameters for reproducibility are listed in the Appendix.

[3] In patient similarity search, we freeze BEHRT and output the learned feature directly.
[4] We use the supervised NCL-$n_y$ since the unsupervised version NCL-$n_w$ requires hourly based sequence, which is not applicable here.
[5] Since the downstream task classifier for linear evaluation of clinical tasks on GKI uses Logistic Regression, we keep it the same here.
[6] GIN-0 is used since it is more powerful than GIN-$\epsilon$ in the original paper.
[7] We follow the same node embedding size (512) as in https://github.com/kavehhassani/mvgrl.

## 5 RESULTS

### 5.1 Clinical evaluations

***Our method outperforms state-of-the-art without supervised pre-training and task-specific fine-tuning on each disease across different tasks:***

Table 1. Performance comparison on treatment outcome prediction task under AUROC and F1-Macro in percentages. All scores are averaged over 5 times repeated 10-fold cross-validation with their standard deviation reported. Scores with boldface denote the best result. In each baseline group, the underline scores stand for the best in the group. * indicating statistical significance over baselines with $p$-value $\leq 0.005$.

| | **Diabetes** | | **Hypertension** | | **Hyperlipidemia** | |
|---|---|---|---|---|---|---|
| Model | AUROC | F1-Macro | AUROC | F1-Macro | AUROC | F1-Macro |
| Retain | 49.6 ± 0.2 | <u>49.0 ± 0.1</u> | 50.0 ± 0.3 | 49.0 ± 0.1 | <u>50.0 ± 0.2</u> | 49.2 ± 0.1 |
| Dipole | <u>50.0 ± 0.2</u> | <u>49.0 ± 0.1</u> | <u>50.0 ± 0.2</u> | <u>50.0 ± 0.1</u> | 49.9 ± 0.2 | <u>49.4 ± 0.1</u> |
| GCT | 49.6 ± 0.3 | 47.9 ± 0.0 | <u>50.0 ± 0.2</u> | 48.2 ± 0.1 | 50.4 ± 0.3 | 48.3 ± 0.1 |
| DeepPatient | 57.9 ± 0.2 | 47.7 ± 0.0 | 58.2 ± 0.3 | 48.1 ± 0.0 | 58.9 ± 0.2 | 48.0 ± 0.0 |
| Seq2SeqAE | 59.5 ± 0.3 | 48.9 ± 0.1 | 59.9 ± 0.2 | 48.4 ± 0.1 | 60.1 ± 0.2 | 48.5 ± 0.1 |
| BEHRT | <u>61.6 ± 0.3</u> | <u>53.0 ± 0.2</u> | <u>62.9 ± 0.2</u> | <u>51.6 ± 0.2</u> | <u>61.4 ± 0.3</u> | <u>51.8 ± 0.2</u> |
| SimCLR | 48.4 ± 0.2 | 51.1 ± 0.1 | 48.0 ± 0.2 | 50.7 ± 0.1 | 48.3 ± 0.2 | 50.4 ± 0.1 |
| SCL | 55.3 ± 0.3 | 52.2 ± 0.2 | 55.1 ± 0.2 | 47.0 ± 0.1 | 52.7 ± 0.3 | 46.5 ± 0.1 |
| NCL | <u>57.0 ± 0.2</u> | <u>53.0 ± 0.1</u> | <u>55.7 ± 0.3</u> | <u>52.9 ± 0.1</u> | <u>53.8 ± 0.3</u> | <u>52.9 ± 0.2</u> |
| InfoGraph | <u>67.5 ± 0.1</u> | <u>53.0 ± 0.1</u> | 66.8 ± 0.2 | <u>51.6 ± 0.1</u> | 67.2 ± 0.1 | <u>51.8 ± 0.1</u> |
| GraphCL | 65.5 ± 0.2 | 51.4 ± 0.1 | 66.2 ± 0.2 | 49.1 ± 0.1 | 65.4 ± 0.2 | 50.3 ± 0.1 |
| MVGRL | 67.1 ± 0.2 | 49.8 ± 0.1 | <u>67.8 ± 0.1</u> | 49.0 ± 0.1 | <u>67.8 ± 0.2</u> | 48.8 ± 0.1 |
| Logistic Regression | <u>57.7 ± 0.2</u> | <u>53.4 ± 0.2</u> | <u>57.5 ± 0.2</u> | <u>52.8 ± 0.1</u> | <u>57.2 ± 0.2</u> | <u>52.9 ± 0.1</u> |
| Random Forest | 65.3 ± 0.2 | 47.7 ± 0.0 | 65.0 ± 0.2 | 48.2 ± 0.0 | 64.7 ± 0.2 | 48.0 ± 0.0 |
| Gradient-boosted trees | 65.2 ± 0.3 | 48.6 ± 0.1 | 66.0 ± 0.1 | 48.5 ± 0.1 | 65.9 ± 0.2 | 48.4 ± 0.0 |
| **GKI** | **68.7 ± 0.2*** | **54.8 ± 0.1*** | **68.3 ± 0.1*** | **53.5 ± 0.1*** | **68.3 ± 0.1*** | **53.6 ± 0.1*** |

We first investigate whether the learned representations from Graph Kernel Infomax provide superior performance on all tasks without supervised fine-tuning. This also mimics the real world such that most clinical decision problems are unknown in advance. Table 1 and 2 show that our proposed GKI consistently outperforms all baseline approaches with statistical significance on treatment plan mortality prediction (Table 1) and patient similarity search (Table 2) for all chronic diseases without supervised pre-training and fine-tuning. Compared to Supervised Learning baselines, where they all fail to deliver good results, we obtain higher scores without label supervision. Surprisingly, GCT does not provide promising results, where we assume the imbalanced distribution may generate similar hidden EHR structures shared across the majority class. In other words, the training signal from the label itself is somehow biased on complex EHR as depicted in Figure 1, which harms the representation learning.

Table 2. Performance comparison on patient similarity search under Precision@1 (P@1), and Precision@10 (P@10) in percentages. All scores are averaged over 5 times repeated 10-fold cross-validation with their standard deviation reported. Scores with boldface denote the best result. In each baseline group, the underline scores stand for the best in the group. * indicating statistical significance over baselines with $p$-value $\leq 0.005$.

| | **Patient Similarity Search** | |
|---|---|---|
| Model | P@1 | P@10 |
| DeepPatient | 0 ± 0 | <u>35.7 ± 0.0</u> |
| Seq2SeqAE | 0 ± 0 | 35.6 ± 0.0 |
| BEHRT | <u>14.8 ± 0.1</u> | 33.8 ± 0.0 |
| SimCLR | <u>22.3 ± 0.1</u> | <u>35.0 ± 0.0</u> |
| SCL | 6.0 ± 0.0 | 20.7 ± 0.0 |
| NCL | 6.0 ± 0.0 | 21.1 ± 0.0 |
| InfoGraph | 15.4 ± 4.0 | 35.3 ± 0.0 |
| GraphCL | <u>21.3 ± 0.9</u> | 35.1 ± 0.0 |
| MVGRL | 19.8 ± 6.0 | <u>35.9 ± 0.0</u> |
| **GKI** | **31.2 ± 0.1*** | **38.4 ± 0.2*** |

In Unsupervised baselines, BEHRT demonstrates superior BERT-based pre-training ability, surpassing all Supervised baselines and Unsupervised baselines. However, its performance on patient similarity search drops when freezing the model. It suggests the necessity of fine-tuning on downstream tasks. For autoencoder-based baselines, DeepPatient and Seq2SeqAE achieve higher Precision@10 scores than BEHRT on patient similarity search, yet are unable to provide good performance on a treatment outcome prediction task. On the contrary, our method achieves preferable results on all tasks with only a single pretrained model.

The performance gain on (Graph) Contrastive Learning methods is inconsistent. SimCLR obtains the highest precision scores in Contrastive baselines on patient similarity search, however, it performs worse than others on treatment plan mortality prediction. NCL receives the competitive Macro-F1 on prediction task compared to all other baselines; even so, the AUROC is not comparable to Graph Contrastive Learning baselines, nor SCL. This also aligns with our claim above; the problematic label supervision training signal. The same situation happens in Graph Contrastive Learning, not a single method can achieve the best results. Notably, InfoGraph displays a consistent competitive performance on all tasks. Recall Section 3.4.1 and Figure 1, where EHR complexity yields sub-optimal data augmentation and label supervision on learned representation for downstream tasks. InfoGraph does not perform augmentation by transforming graph topology and node features which is validated in our evaluation.

Finally, we validate whether the level of performance obtained by GKI in clinical applications confirms state-of-the-art as long as simpler alternatives, e.g., classic machine learning models, are potentially achievable. The GKI outperforms all traditional baselines, e.g., Logistic Regression, Random Forest, and Gradient-boosted Tree. Compared to supervised training on classic models, the "frozen" representation produced by GKI sufficiently provides consistent performance gain without disease-specific pre-training or fine-tuning. It also verifies the effectiveness and potential of self-supervised learning on large-scale EHRs.

### 5.2 Augmentation scheme evaluations

***Our proposed augmentation scheme works with different graph encoders:***

Next, we examine the performance behavior of our proposed Kernel Subspace Augmentation on different graph encoder backbones. In practice, it is common to select different graph encoders to gain better performance on different clinical tasks or diseases. Moreover, the contrastive framework should maintain robustness, which constantly provides performance boosts for selected different types of graph encoders. Figure 7 reports the averaged Macro-F1 and Precision@10 of different GNNs with our proposed augmentation framework under clinical evaluation tasks. We can see that GKI maintains more increased performance across all types of GNNs. This further suggests the effectiveness of Kernel Subspace Augmentation which is capable of providing robust augmented views for self-supervised graph representation learning on highly noisy patient EHRs. On the other hand, we notice that GAT provides the best overall performance in clinical evaluation tasks, yet GCN can not produce better results. It is reasonable that the attention mechanism in GAT captures more complex hidden relationships between medical events, as compared to the simple GCN when learning the self-distinguished features from different kernel subspace augmented views.

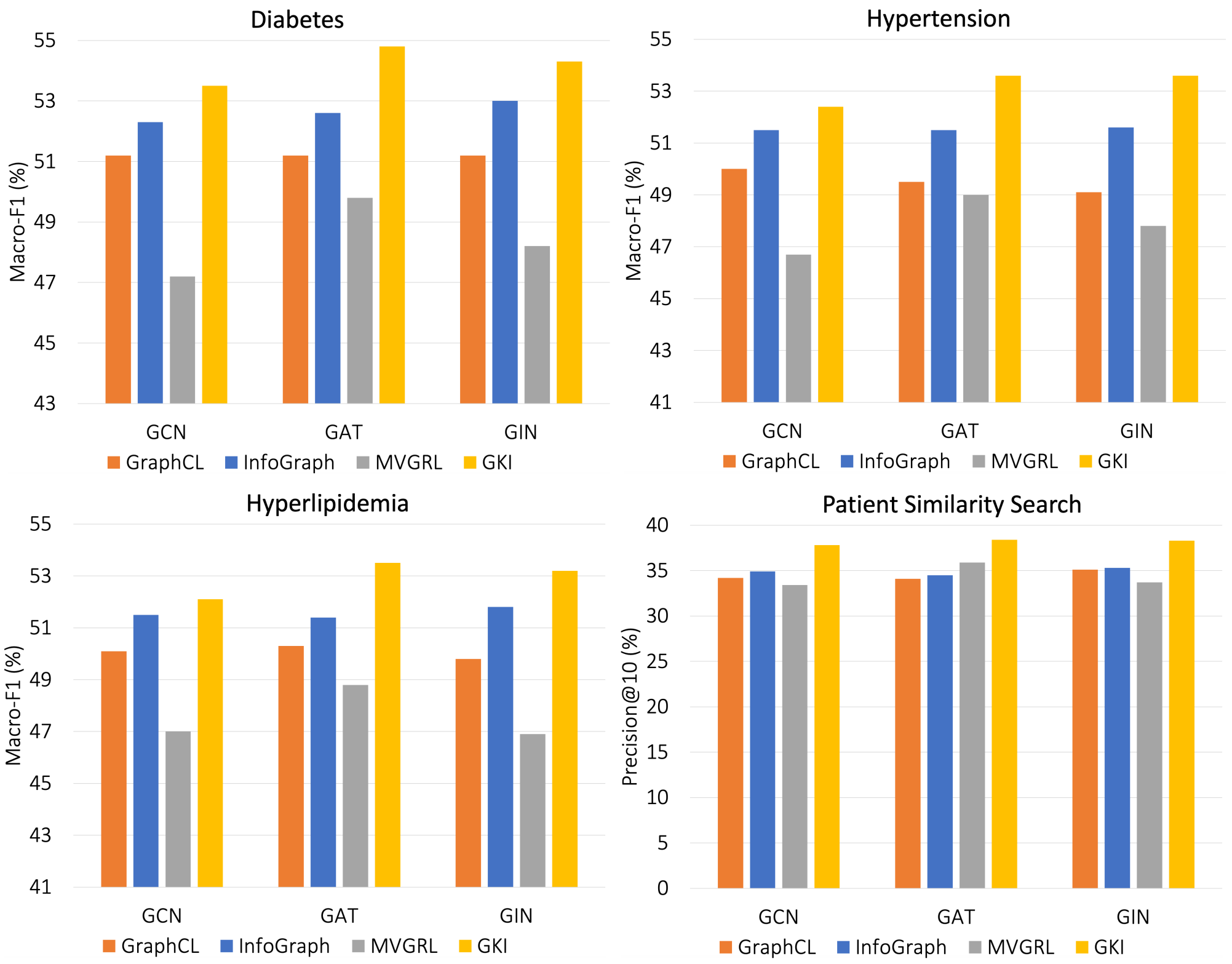

Fig. 7. Macro F1 and Precision@10 (all in %) on different graph encoders (GNNs) under clinical evaluation tasks. GKI maintains the best performance under different types of GNN compared with other self-supervised graph representation learning methods, showing the robustness of our proposed augmentation scheme.

### 5.3 Non-ad hoc task evaluations

***Our proposed framework provides non-ad hoc generalizability on non-clinical domains:***

Table 3. Performance comparison on graph classification benchmarks under linear evaluation protocol on averaged accuracy in % with their standard deviation reported. The boldface denotes the best result. GKI outperforms the state-of-the-art baselines in 6 out of 8 benchmarks. * indicating statistical significance over baselines with $p$-value $\leq 0.005$. For all baselines, we report the evaluation scores from their original paper directly.

| | PTC-MR | PROTEINS | NCI1 | COLLAB | IMDB-B | IMDB-M | RDT-B | RDT-M5K |
|---|---|---|---|---|---|---|---|---|
| InfoGraph | 61.7 ± 1.4 | 74.4 ± 0.3 | 76.2 ± 1.4 | 70.7 ± 1.1 | 73.0 ± 0.9 | 49.7 ± 0.5 | 82.5 ± 1.4 | 53.5 ± 1.0 |
| GCC | - | - | - | 79.4 | **75.6** | 50.9 | 87.8 | 53.0 |
| GraphCL | 61.3 ± 2.1 | 74.4 ± 0.5 | 77.9 ± 0.4 | 71.4 ± 1.2 | 71.1 ± 0.4 | 49.2 ± 0.6 | **89.5 ± 0.8** | 56.0 ± 0.3 |
| MVGRL | 62.5 ± 1.7 | - | 77.0 ± 0.8 | 76.0 ± 1.2 | 74.2 ± 0.7 | 51.2 ± 0.5 | 84.5 ± 0.6 | - |
| **GKI** | **64.0 ± 0.8*** | **76.5 ± 0.4*** | **79.3 ± 1.9*** | **80.4 ± 0.1*** | 74.5 ± 0.5 | **51.6 ± 0.4** | 86.8 ± 0.2 | **56.0 ± 0.1** |

Table 4. Dataset statistics for efficiency evaluation. For graph classification, we select representative datasets with different scales for efficiency benchmarks.

| | MIMIC-IV | COLLAB | RDT-B | RDT-M5K |
|---|---|---|---|---|
| # graphs | 230,312 | 5,000 | 2,000 | 4,999 |
| Max # nodes | 5,637 | 492 | 3,782 | 9,566 |
| Max # edges | 188,844 | 80,238 | 8,142 | 3,648 |
| Avg # nodes | 66.5 | 74.5 | 429.6 | 508.5 |
| Avg # edges | 1,015.2 | 2,457.8 | 497.8 | 594.9 |

To further bridge our proposed method to other domains, we evaluate Graph Kernel Infomax on graph classification benchmarks to see if it is an ad hoc learning framework in the clinical domain. Contrasting Table 1 and Table 3, Graph Kernel Infomax not only surpasses all the baselines on both complex clinical prediction and patient similarity search tasks but also can generalize well to other graph classification tasks. In Table 3, our method outperforms state-of-the-art graph contrastive learning frameworks in 6 out of 8 benchmarks. For lower scores, IMDB-B is the second highest, and RDT-B is still higher than MVGRL and InfoGraph. For results that are not statistically significant, e.g., IMDB-M and RDT-M5K, our method obtains competitive or even equivalent performance against the highest SOTAs. This gives us an insight that Graph Kernel Infomax performs better than those contrastive learning counterparts on clinical applications with the same ability on graph classification benchmarks. This further suggests the generalizability of solving unexplored problems without any prior knowledge.

## 6 DISCUSSION

### 6.1 Efficiency analysis

When training on large-scale graphs, one often faces computation challenges regarding time spent and memory usage as the number of nodes and edges significantly increases. We evaluate the running time and memory footprint of the MIMIC-IV patient graph pre-training of GKI against SOTA self-supervised graph representation learning. To evaluate the efficiency on different scales of graphs, we select three different sizes of graph classification benchmark datasets listed in Table 4. We notice that hyperparameters used by MVGRL[8] lead to a severe OOM (Out-of-Memory) problem. For a fair comparison, we fix the graph encoder as a two-layer GCN[9] with the batch size selected as the "largest runnable"[10] among all methods. For GKI, we choose 128 as landmark points $k$.[11].

Table 5 demonstrates that GKI provides highly efficient pre-training on the lowest memory usage and competitive running time compared with all SOTA methods. Specifically, given the importance of how memory usage directly restricts the scalability of model training, GKI maintains its excellent memory efficiency on different scales of graphs in terms of different-scale nodes or edges, indicating the prominence of our core idea: constructing augmented views without sophisticated operations, leading to memory-saving pre-training.

***GKI overcomes computational challenges when dealing with large datasets by the following:***

- GKI does not perform costly operations to create augmented views, e.g., changing structures or costly diffusion on graphs described in GraphCL and MVGRL. On the contrary, our proposed Kernel Subspace Augmentation only introduces a limited amount of parameters, e.g., landmark points, to create perturbation on different graph views in kernel subspace.

[8] Node embedding dimension set to 512. Note that, the evaluation scores in Table 5.3 are directly cited from the original paper of baselines.
[9] In order to keep the evaluation less biased and sensitive due to additional GNN components, e.g., attention heads or isomorphisms, we select the simplest architecture: GCN as the backbone graph encoder.
[10] The "largest runnable" means the batch size that can fit the GPU memory for training all self-supervised graph representation learning baselines.
[11] We keep the landmark point $k$ with the same setting of the graph classification benchmark described in Section 4.3.

Table 5. Memory usage and running time comparison of pertaining task on different scales datasets (see Table 4). Memory is measured at peak usage in megabytes (MB) in peak usage as it is a direct indicator of scalability. All evaluated running times are measured in seconds (s) for a single epoch training. OOM means Out of Memory error which training terminates before complete. Our evaluation is performed on a workstation with an i9-10900X 3.7GHz CPU, 256GB DDR4-2400 RAM, and a single Nvidia RTX A6000 48GB GPU.

| | **MIMIC-IV** | **COLLAB** | **RDT-B** | **RDT-M5K** |
|---|---|---|---|---|
| Selected batch size | 128 | 32 | 8 | 32 |
| GraphCL | 26.96s / 620 MB | 38.87s / 27999 MB | 16.43s / 5087 MB | 45.04s / 9935 MB |
| InfoGraph | 40.41s / 3120 MB | 16.46s / 27792 MB | 14.64s / 5905 MB | 39.85s / 18896 MB |
| MVGRL | 474.92s / 5240 MB | 13.17s / 5373 MB | 34.86s / 30505 MB | OOM > 48 GB |
| GKI | 60.45s / **559 MB** | 9.31s / **2743 MB** | 18.2s / **2105 MB** | 38.46s / **6153 MB** |

- GKI does not create dual-branch representation on augmented graph views for contrastive learning. Instead, ours performs contrastive learning on approximated manifolds yielded by different geometries. In such a case, we avoid dual-graph representation learning that grows memory consumption and thus restricts usage by a limited amount of landmark points.
- GKI does not contrast each graph with negative samples from other graphs in the same training batch. On the contrary, ours only contrasts them with itself without the need to compare with others (refer to Equation 21, 22, and 23). This results in a smaller batch size needed to train the model.

### 6.2 The choice of $k$ landmark points

Since our augmentation scheme is parameterized by the number of landmarks $k$ selected that affect the clustering performance, choosing the right $k$ is thus important. We analyze how our proposed framework is sensitive to the number of landmark points selected. Figure 8 reports the averaged Macro-F1 and Precision@10 of our augmentation framework under different GNNs with different $k$ selections on clinical evaluation tasks. In general, for the number of landmarks $k$, as $k$ increases, the result improves especially on GAT. As discussed in the previous section, GAT tends to obtain better performance as its attention mechanism can seize more information from the representation space. When landmark points increase, more cluster groups can provide distinguished feature information for GAT to learn. However, it is not sensitive to $k$ when performing on patient similarity search. We can see that the precision@10 scores are comparable even when $k$ is small indicating our learning framework can extract as much structural information as possible to sufficiently provide evidence on similarity measure.

### 6.3 The effect on pseudo-inverse setting to $I$

As described in Section 3.3.3, the performance is improved by explicitly setting the pseudo-inverse to $I$ when constructing augmented views through kernel feature map. This is identical to skipping the Singular Value Decomposition (SVD) step when computing the pseudo-inverse. From Figure 9, we can notice the performance drop on Macro-F1 on both evaluation scores and convergence when performing SVD during the contrastive learning process. We surmise the reason is that SVD defeats the learning signals to the clustering layer by zeroing out with the singular values associated with each landmark point. Moreover, setting pseudo-inverse to $I$ implicitly defines an orthogonal constraint, since the inverse of the orthogonal matrix is identity. We leave it here as future research.

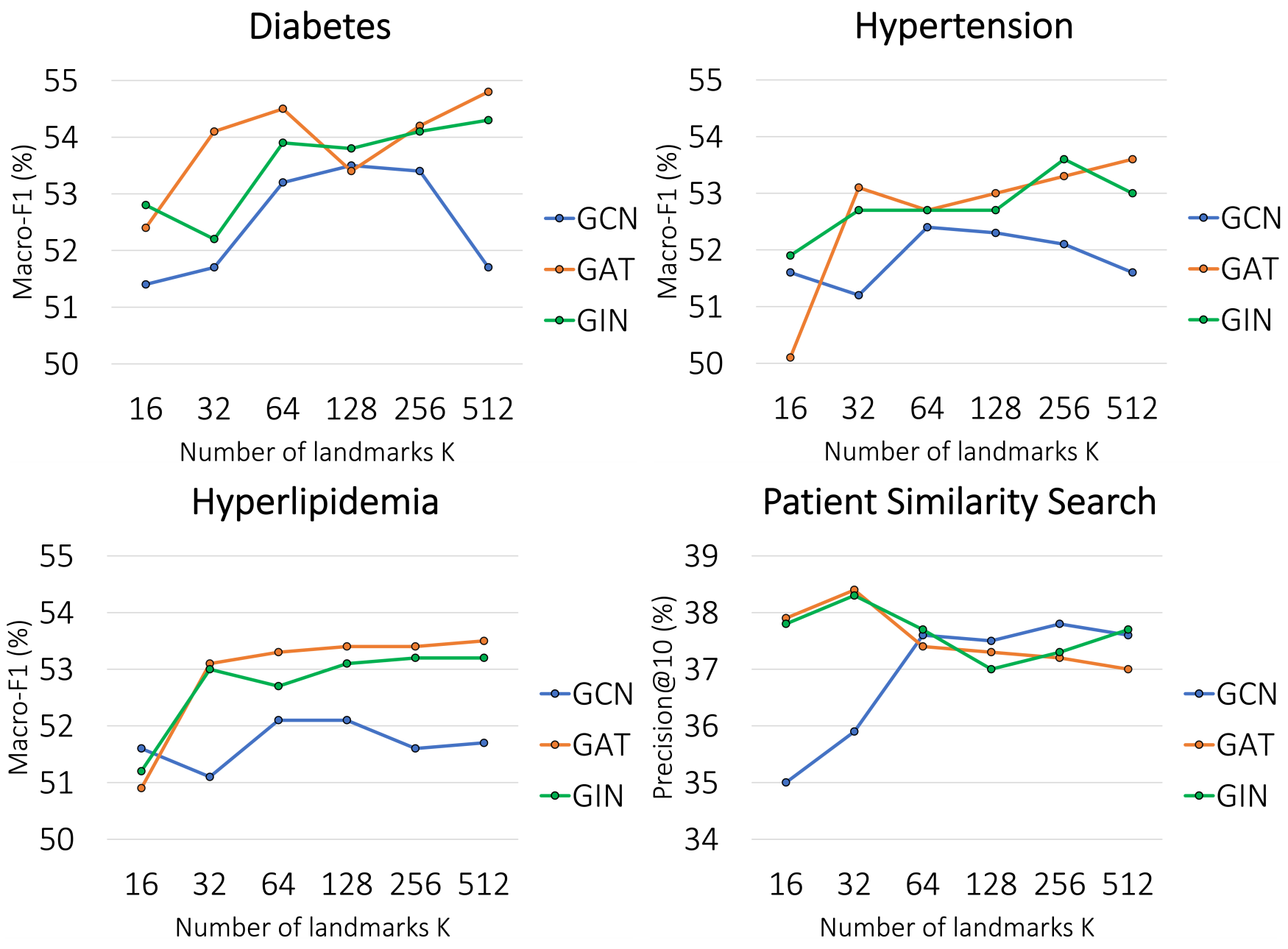

Fig. 8. Macro F1 and Precision@10 (all in %) on different graph encoders (GNNs) with different numbers of landmarks $K$.

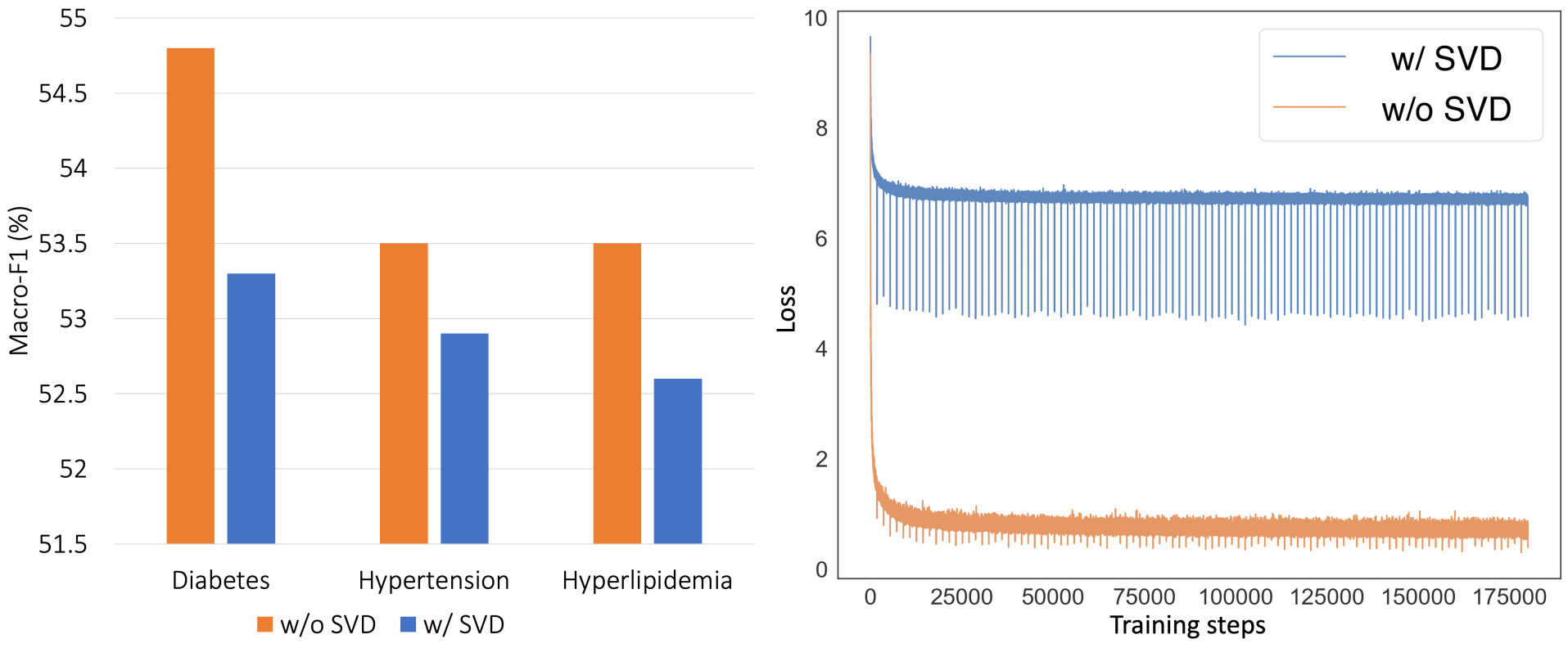

Fig. 9. Comparison of training with and without SVD.

### 6.4 Error analysis

The primary functionality of GKI is to provide self-discriminative features for patient representations, where error occurs when their discriminative power is lacking for downstream task models. We investigate how GKI makes errors in the patient similarity search since it is apparent to contrast two different cohorts but highly similar patients that easily direct wrong top-K similarity search results. Figure 10 shows three hypertension patients (A, B, and C) where the middle denotes the query patient B whose hypertension treatment has failed, and both left and right are the returned top two similar patients A and C[12] whose treatment plans are successful. We can easily see that patients A and B are nearly identical except for different thicknesses of edges around the gender node and between the first and second visit[13]. Consequently, GKI produces nearly equivalent patient representations without identifying temporal discrepancy, e.g., different ages and visit times, given that two patients are in different treatment conditions. On the other hand, patient C only has a one-time visit where the diagnosis and drug prescription are similar to the first visit in patient B. In this case, GKI produces similar representations without comprehending their different visit history. These observations reveal that GKI might not well-capture temporal information during self-supervised contrastive learning, thus delivering ambiguous self-discriminative features leading to erroneous downstream task results. We surmise that incorporating explicit edge attributes or temporal relations into our kernel subspace augmentation instead of naive weighted edges resolves the issue.

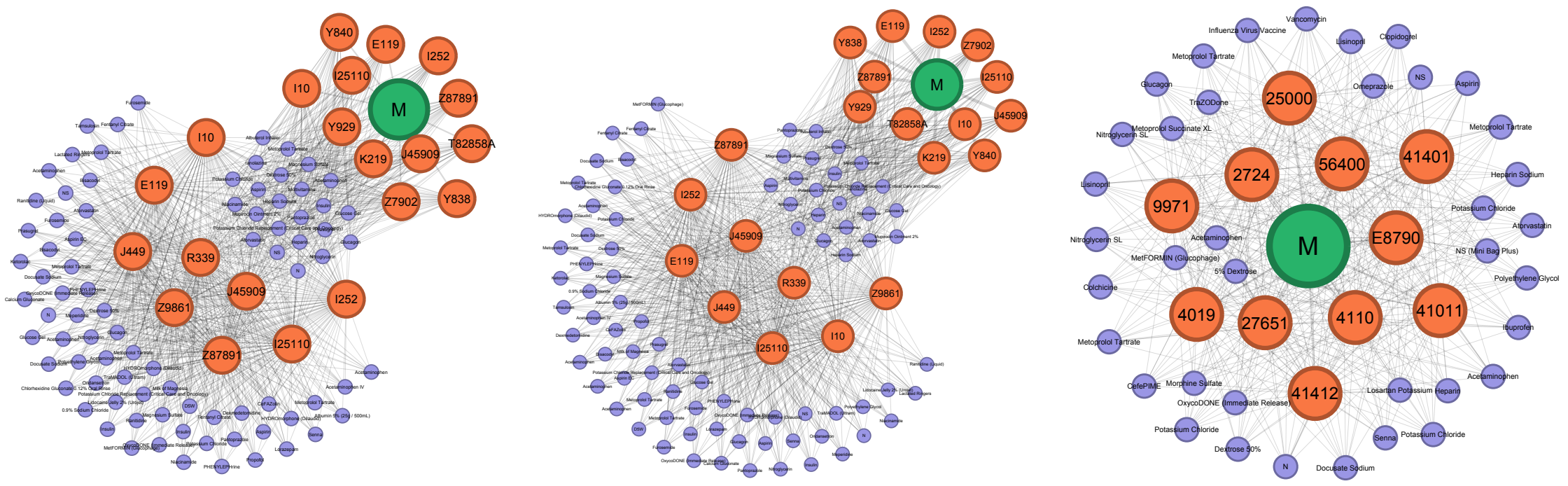

Fig. 10. Visualizations of three hypertension patient graphs (A, B, and C) where the left (A) and the right (C) are the top two similar to the middle (B). The middle graph shows a hypertension patient with failed treatment. Both left and right of the middle are hypertension patients with successful treatment. The colors green, orange, and purple denote patient gender, disease diagnosis, and drug prescriptions respectively. The darkness of the edge indicates a larger edge weight, e.g., long time periods or older ages.

## 7 LIMITATION AND FUTURE WORK

In addition to the observed deficiency and potential remedy of GKI in capturing miniature temporal information discussed in the previous section, we would like to point out additional limitations of our proposed method. Empirically, GKI demonstrates how graph-based self-supervised learning on patient EHRs benefits clinical tasks. However, real-world EHR consists of multi-modality heterogeneous features such as clinical notes, lab tests, or even medical images. Although our theoretical framework can work directly with graphs that contain heterogeneous nodes, the proposed EHR graph formulation only considers homogeneous nodes, e.g., text-based medical code, instead of heterogeneous ones and, thus, diminishing utility in such scenarios. Another problem regards the graph complexity due to the longer medical history that results in an extremely long chain patient graph, e.g., Figure 14 in the Appendix. Such an issue creates limitations on modeling very long-term temporality since Graph Neural Networks (GNN) often ill-perform when stacking deeper layers in order to capture more in-depth information [48]. Consequently, future work directions include but are not limited to "multi-modality" heterogeneous graph-based modeling, more simple and compressed EHR graph formulation, and miniature temporal information capturing through self-supervised learning.

[12]Here, A is more similar to B than C.
[13]As the graph is too small to see, we recommend using the PDF version to zoom in and see the thickness difference.

## 8 CONCLUSION

The complexity of EHR challenges patient representation learning. Many existing efforts addressing such hardship fail to deliver preferable predictive accuracy nor generate highly transferable patient representations for various clinical applications. Consequently, our proposed Graph Kernel Infomax not only overcomes the EHR augmentation problem but also provides a simple and powerful contrastive learning paradigm. We achieve state-of-the-art performance on both clinical application and graph classification benchmarks. Potentially, the highly efficient training of GKI allows larger-scale graph-based "multi-modality" medical pre-trained model development. Technically, our approach can likely apply to other domains when graph formulation is determined. Practically, the discussed approach is under limited clinical use and assessment by medical clinicians.

## A TRAINING ALGORITHM FOR GRAPH KERNEL INFOMAX

**Algorithm 1** GKI training loop in PyTorch-like pseudocode

```
# GNN: A L-layer Graph Neural Network as Graph encoder
# k_means: L independent k-means clustering
# Kernel_E: Euclidean distance kernel feature map
# Kernel_S: Spherical distance kernel feature map
# SUM: Sum pooling on nodes to derive graph
# CONCAT: Concatenation operator
# Proj_node: Node projection head
# Proj_graph: Graph projection head
# optimizer: SGD optimizer

for G in batch: # Load a mini-batch from all patient graphs
    X, A = G # Node features X and adjacency matrix A
    # Obtain node representation from all layers
    node_reps = GNN(X, A)
    L_rec = 0
    graph_Es = [] # Euclidean kernel feature for graph from each layer
    graph_Ss = [] # Spherical kernel feature for graph from each layer

    for i in L:
        # K-means clustering on node representation for each layer,
        # and construct kernel features for each layer
        K_means[i](node_reps[i])
        L_rec_i = K_means.loss_rec # Get layer-wise clustering loss
        L_rec = L_rec + L_rec_i # Layer wise k-means loss are added
        C = K_means[i].centroids # Obtain cluster centroids
        # Construct Euclidean kernel feature for layer i
        node_E = Kernel_E(node_reps[i], C)
        graph_Es.append(SUM(node_E)) # Derive graph representation
        # Construct Spherical kernel feature for layer i
        node_S = Kernel_S(node_reps[i], C)
        graph_Ss.append(SUM(node_S)) # Derive graph representation

    # Concatenate graph from each layer
    graph_E = CONCAT(graph_Es)
    graph_S = CONCAT(graph_Ss)
    # node_E and node_S are last layer kernel features for node
    # Projection on kernel features for nodes and graph
    p_node_E = Proj_node(node_E)
    p_node_S = Proj_node(node_S)
    p_graph_E = Proj_graph(graph_E)
    p_graph_S = Proj_graph(graph_S)
    # Compute contrastive loss
    L_N_G = Loss_N_G(p_node_E, p_graph_S)
            + Loss_N_G(p_node_S, p_graph_E)
    L_G_G = Loss_G_G(p_graph_E, p_graph_S)
            + Loss_G_G(p_graph_S, p_graph_E)
    L = L_N_G + L_G_G + L_rec
    # Compute gradients and update parameters
    L.backward()
    optimizer.step()
```

Table 6. Detailed hyperparameters for experiments.

| | Clinical Task | Graph Classification |
|---|---|---|
| Graph Encoder $f_\theta$ | {GCN, GAT, GIN-0} | GCN |
| # $f_\theta$ layers | 2 | $\{1, 2, 3, 4\}$ |
| $f_\theta$ hidden sizes | 128 | 512 |
| Activation for $f_\theta$ | PReLU | PReLU. |
| Number of clusters $k$ | 512 | 128 |
| $f_\alpha$ hidden sizes | (128, 128, 128) | (128, 128, 128) |
| # $f_\alpha$ layers | 3 | 3 |
| Activation for $f_\alpha$ | PReLU | PReLU. |
| # $f_\beta$ layers | 3 | 3 |
| $f_\beta$ hidden sizes | (128, 128, 128) | (128, 128, 128) |
| Activation for $f_\beta$ | PReLU | PReLU. |
| Similarity measure | Cosine | Cosine |
| Temperature | 0.01 | {0.1, 0.01, 0.001} |
| Epochs | 100 | {10, 20, 40, 100} |
| Batch size | 128 | {32, 64, 128, 256} |
| Optimizer | Adam | Adam |
| Learning rate | 0.001 | 0.001 |
| Final output sizes | $512 \times 2$ | $128 \times \{1, 2, 3, 4\}$ |

Table 7. Data statistics for MIMIC-IV dataset

| | MIMIC-IV-Total |
|---|---|
| # patients | 230,312 |
| Max # nodes | 5,637 |
| Max # edges | 188,844 |
| Avg # nodes | 66.5 |
| Avg # edges | 1,015.2 |

## B IMPLEMENTATION DETAILS

Our method is implemented using PyTorch 1.10 [29] and PyTorch Geometric 2.0.2 [10]. The source code is available [14] for the review process. Table 6 shows the hyperparameter settings for our model. The grid search is performed to choose the best hyperparameters for each experiment.

## C DATASET

The statistics for MIMIC-IV used for pre-training is summarized in Table 7, and the selected chronic diseases for treatment plan mortality prediction are summarized in Table 8. For patient similarity search, the dataset is the same as treatment plan mortality prediction except the ground-truth label becomes the disease instead of success or failure. Table 9 provides data statistics on graph classification benchmark datasets.

[14] https://anonymous.4open.science/r/Graph-Kernel-Infomax-48E9

Table 8. Data statistics for selected chronic diseases

| | Diabetes | Hypertension | Hyperlipidemia |
|---|---|---|---|
| ICD codes | 250*, E11*, E10* | 401*, I10 | 272*, E78*, E881 |
| # patients | 13,571 | 25,004 | 23,204 |
| # Success | 12,355 | 23,133 | 21,391 |
| # Failure | 1,216 | 1,871 | 1,813 |
| Max # nodes | 1,966 | 1,966 | 1,966 |
| Max # edges | 44,780 | 44,780 | 44,780 |
| Avg # nodes | 81.7 | 70.8 | 71.5 |
| Avg # edges | 1,221 | 952.8 | 994.4 |

Table 9. Data statistics for graph classification benchmarks

| | # graphs | Avg # nodes | Avg # edges | # Classes |
|---|---|---|---|---|
| PTC-MR | 344 | 14.3 | 14.7 | 2 |
| PROTEIN | 1,113 | 39.1 | 72.8 | 2 |
| NCI1 | 4,110 | 29.9 | 32.3 | 2 |
| COLLAB | 5,000 | 74.5 | 2,457.8 | 3 |
| IMDB-B | 1,000 | 19.8 | 96.5 | 2 |
| IMDB-M | 1,500 | 13 | 65.9 | 3 |
| RDT-B | 2,000 | 429.6 | 497.8 | 2 |
| RDT-M5K | 4,999 | 508.5 | 594.9 | 3 |

## D MIMIC-IV DATA PREPROCESSING

We select and join these four tables `patients.csv`, `admissions.csv`, `diagnoses_icd.csv`, and `prescriptions.csv` to formulate patient history by patient unique id. Refer MIMIC-IV documentation [15] for detailed schema. Below is the schema of final joined medical history with core attributes used for a single patient as a json format below:

```
{
    "patient_id": str, # patient unique id
    "history": [ # list of clinical visit sorted by admittime
        {
            "hadm_id": str, # Hospital admission id
            "admittime": date, # Admission datetime
            "dischtime": date, # Discharged datetime
            "deathtime": date, # Mortality datetime
            "gender": str, # Patient gender
            "age": int, # Patient age
            "hospital_expire_flag": bool, # Mortality indicator
            "icd_codes": [str], # List of diagnosis in ICD9 or ICD10
            "days": int, # Duration of this medical event
            "drugs": [str] # List of prescription
        },
    ]
}
```

[15] https://physionet.org/content/mimiciv/1.0/

## E PATIENT SELECTION PROCEDURE

The following pseudocode snippet determines the success or failure label for a given `patient`:

```
for visit in patient["history"]:
    if visit["hospital_expire_flag"] == True:
        label = 1
        break
    label = 0
```

## F ALGORITHM FOR PATIENT GRAPH CONSTRUCTION

**Algorithm 2** Patient graph creation in PyTorch-like pseudocode

```
# The graph is constructed by a set of edges with the format
# (node_connect_from, node_connect_to, edge_weight)
edge_list = []

# We only use gender and age to create the first node for simplicity
first_node = patient["gender"]
age = patient["age"]

for visit in enumerate(patient["history"]):
    # The first node (demographic node) is connected to each
    # diagnosis node in the first visit with edge weight as age
    if index(visit) == 0:
        for diagnosis in visit["icd_codes"]:
            edge = (first_node, diagnosis, age)
            edge_list.append(edge)
    else:
        # For other visit, the diagnosis nodes are connected
        # to each drug node with edge weight as days

        days = visit["days"]
        for drug in visit["drugs"]:
            for code in visit["icd_codes"]:
                edge = (code, drug, days)
                edge_list.append(edge)

        # For other visit, the diagnosis nodes are connected
        # to each drug node in the prior visit with edge weight
        # as time different between them

        prev_visit = get_prev_visit(visit)
        time_diff = visit["admittime"] - prev_visit["dischtime"]
        for drug_prev in prev_visit["drugs"]:
            for diagnosis in visit["icd_codes"]:
                edge = (drug_prev, diagnosis, time_diff)
                edge_list.append(edge)

        # When admission time in the previous visit is in between
        # the current and the previous of previous visit,
        # then we connect the diagnosis node to each drug node
        # in the prior of the prior visit with edge weight as the
        # time difference between them

        prev_2_visit = get_prev_visit(prev_visit)
        if prev_visit["admittime"] < prev_2_visit["dischtime"]:
            time_diff = visit["admittime"] - prev_2_visit["dischtime"]
            for drug_prev_prev in prev_2_visit["drugs"]:
                for diagnosis in visit["icd_codes"]:
                    edge = (drug_prev, diagnosis, time_diff)
                    edge_list.append(edge)
```

## G SAMPLE PATIENT GRAPHS FROM MIMIC-IV

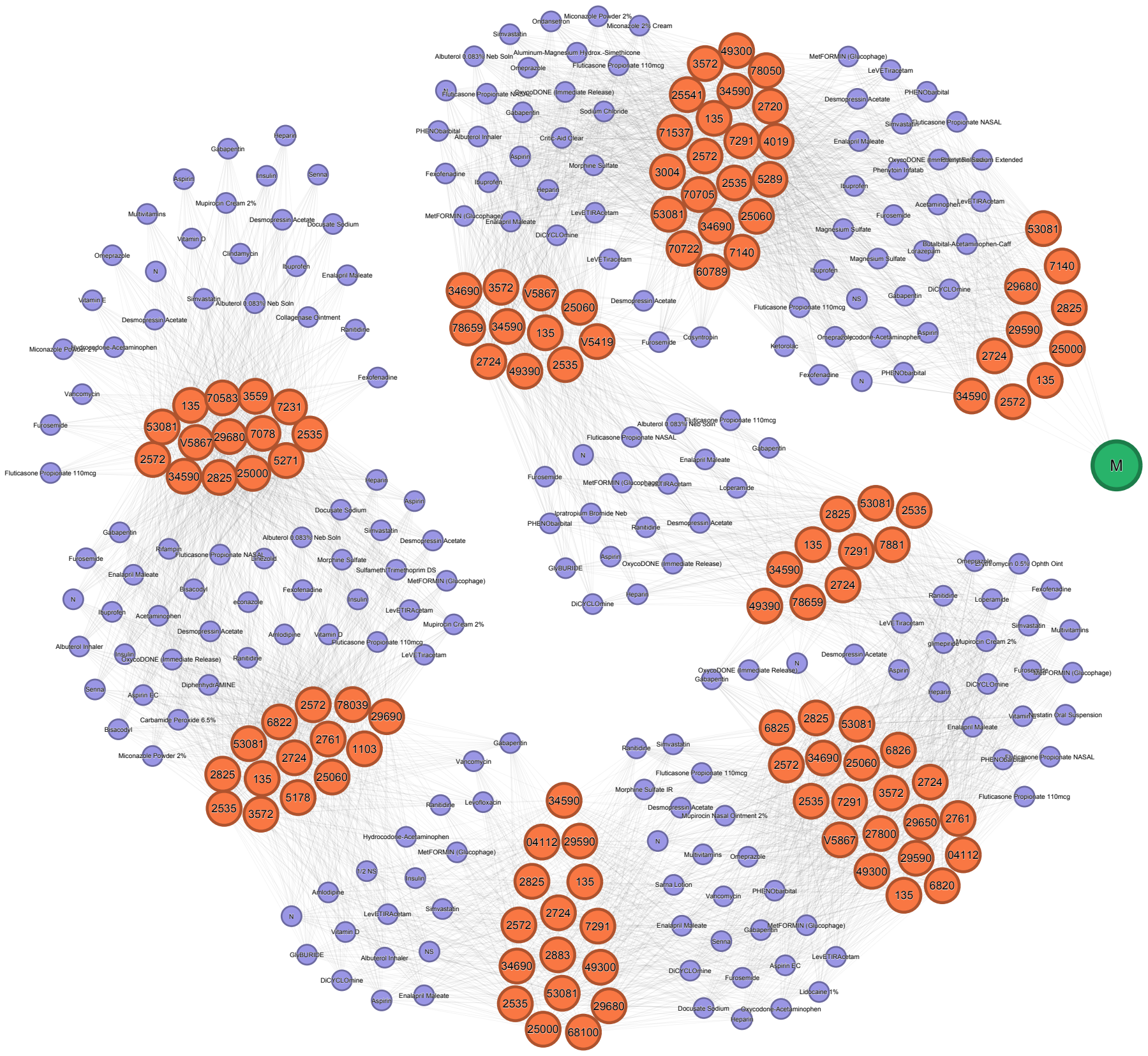

Fig. 11. Visualization of sample patient graph.

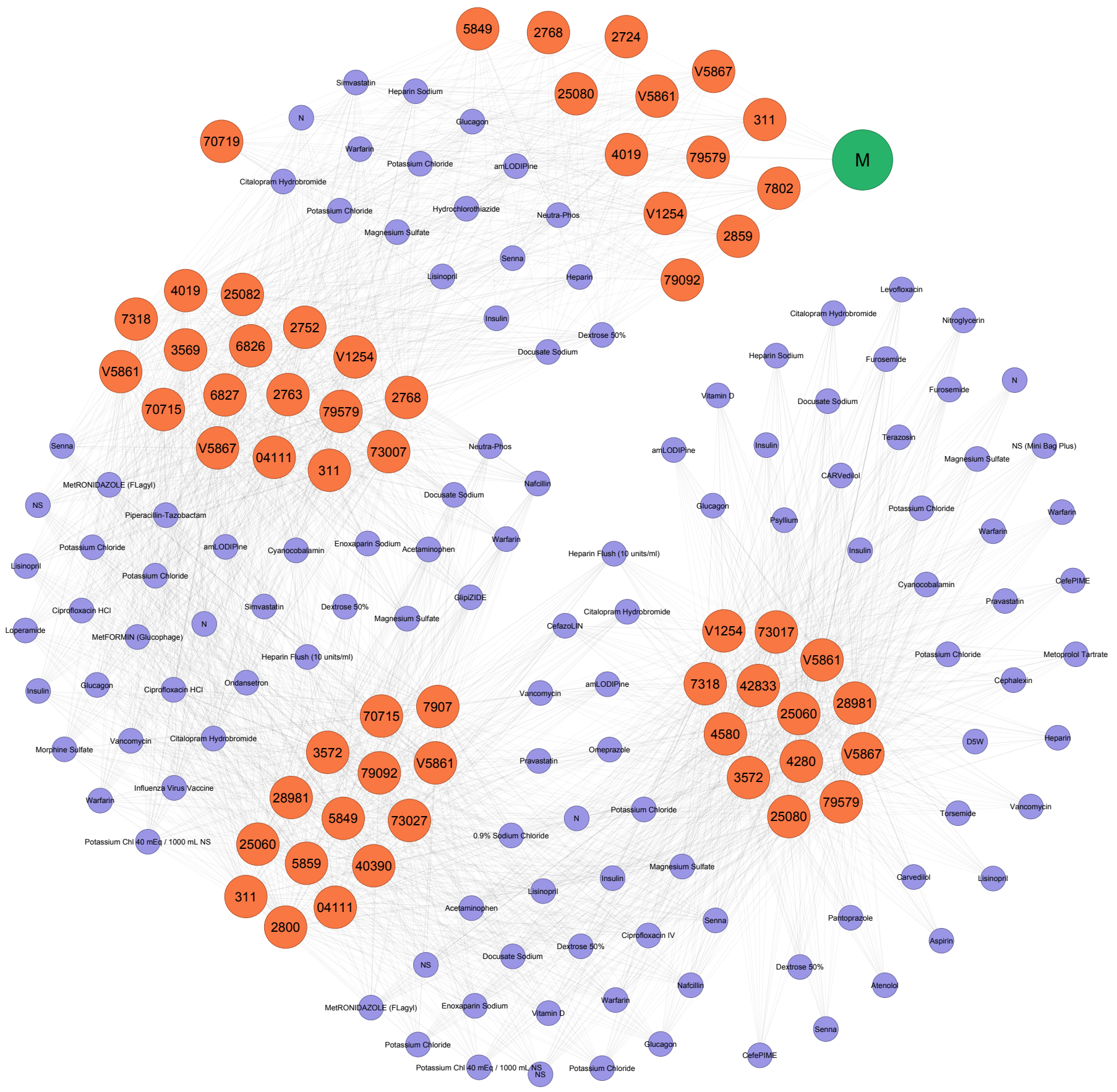

Fig. 12. Visualization of sample patient graph.

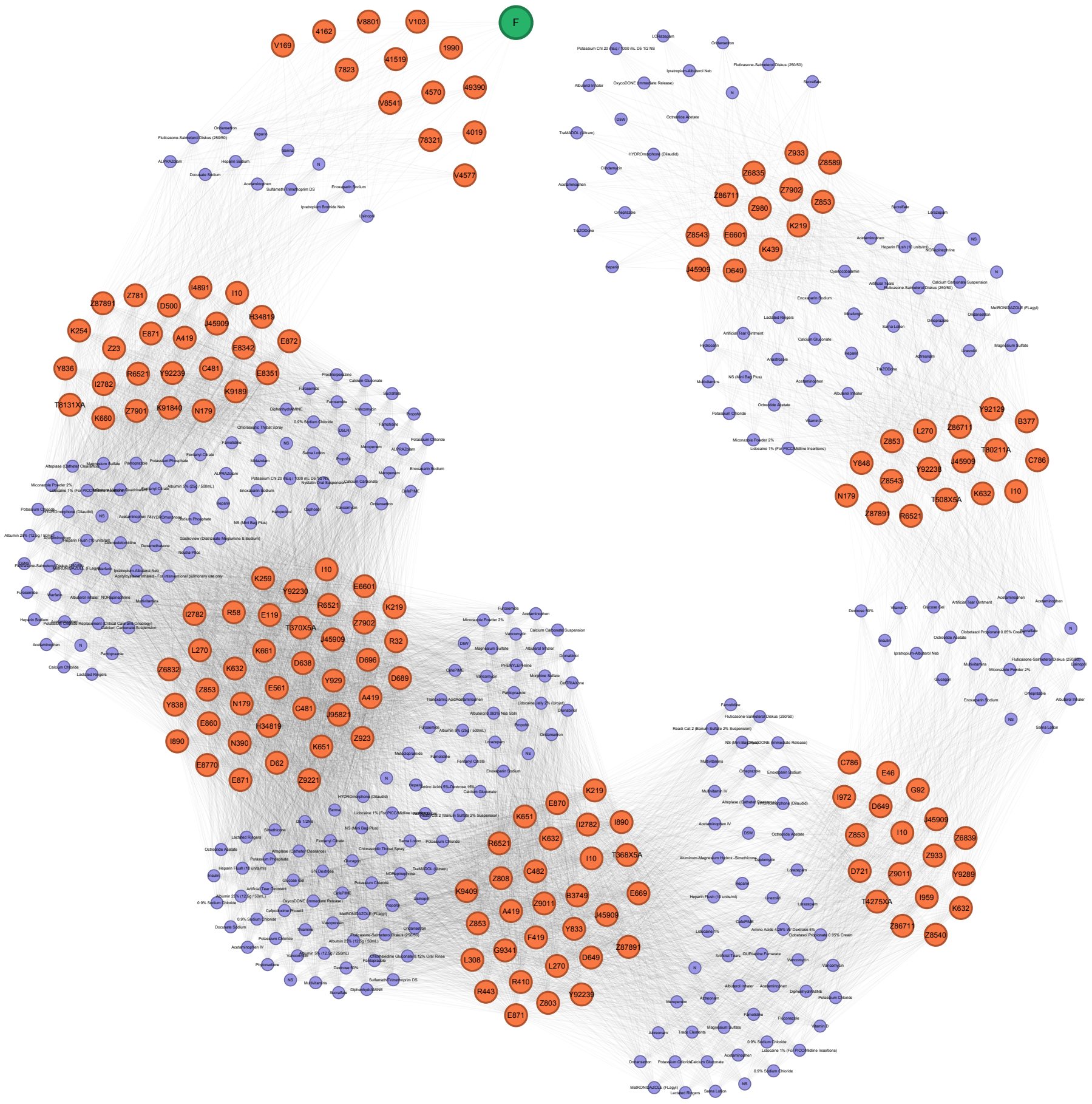

Fig. 13. Visualization of sample patient graph.

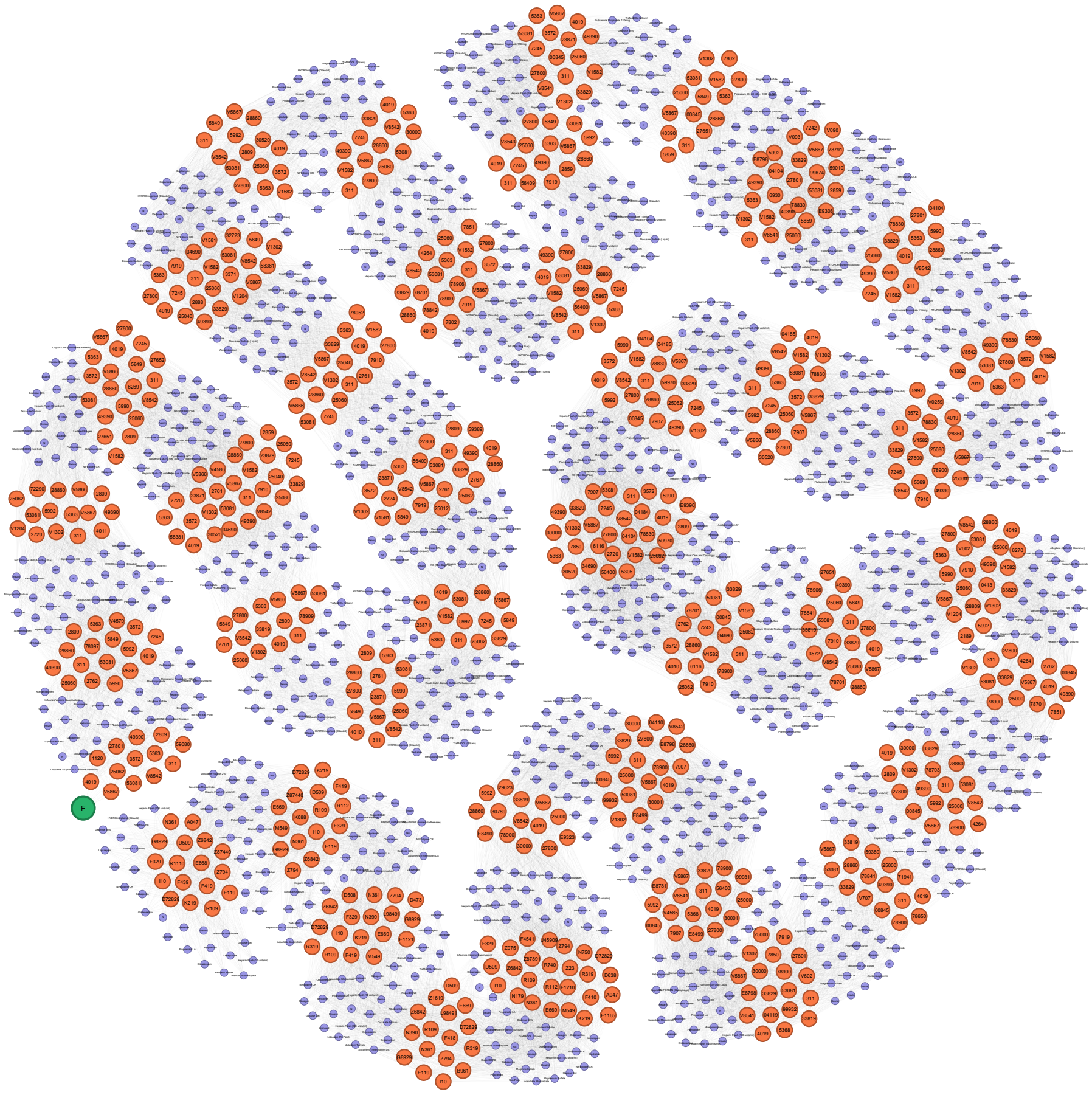

Fig. 14. Visualization of sample long-chain patient graph.